\documentclass[runningheads]{llncs}
\usepackage[T1]{fontenc}
\usepackage{graphicx}
\usepackage{subcaption}
\usepackage{fontspec}
\usepackage{arydshln}
\newfontfamily\arabicfont[ Script=Arabic]{Amiri-Regular.ttf}

\usepackage{booktabs}
\usepackage{array}
\usepackage[normalem]{ulem}

\usepackage{xcolor}
\usepackage{amssymb}
\usepackage{booktabs}
\usepackage{multirow}
\usepackage{amsmath}
\usepackage{graphicx}
\usepackage{booktabs}
\usepackage{array}

\usepackage{makecell}
\usepackage{subcaption}

\begin{document}
\title{Performance Gap Analysis between Latin and Arabic Scripts HTR}
%
%
\author{Sana Al-azzawi\orcidID{0000-0001-7924-4953} \and
Elisa Barney\orcidID{0000-0003-2039-3844} \and
Marcus Liwicki\orcidID{0000-0003-4029-6574}}
\authorrunning{S. Al-azzawi et al.}

\institute{
Luleå University of Technology\\
Department of Computer Science, Electrical and Space Engineering\\
Luleå, 97187, Sweden\\
\email{sana.al-azzawi@ltu.se, elisa.barney@ltu.se, marcus.liwicki@ltu.se}
}
%
%
\date{}
\maketitle              

\begin{abstract}
Recent studies have shown that handwritten text recognition (HTR) systems perform worse on Arabic-script datasets than on Latin-script data. However, the reasons for this gap are still not well understood due to the lack of controlled comparisons. In this work, we present a comprehensive study of Arabic- and Latin-script HTR using a unified CRNN model for line-level HTR across nine datasets (including KHATT (Arabic), Muharaf (Arabic), NUST-UHWR (Urdu), PHTD (Persian), IAM (English), READ-2016 (German), and others) and different training sizes ($K \in \{100, 500, 1000, 2000, \dots, K_{\text{full}}\}$). Our results show the performance gap remains: it is large in low-resource settings, decreases with more data, but remains even at full scale, with a consistent difference of 5–7 CER points. We show that annotation quality matters, as many datasets contain labeling errors. Cleaning reduces error rates and narrows the gap, but does not eliminate it. In addition, we find that a fixed number of training samples provides less effective coverage in Arabic due to higher visual variability, requiring more data to learn similar representations. We compare recognition across datasets in terms of the number of text lines and the number of characters, showing an equivalence trade-off. We compare character frequency distributions across scripts and show that Arabic is significantly more heavy-tailed than Latin. Our error analysis reveals that around 30\% of substitution errors in Arabic datasets (e.g., KHATT) are caused by confusion between visually similar characters, compared to about 15\% in Latin-script datasets such as IAM.

\end{abstract}
\section{Introduction} \label{sec:introduction}
Handwritten Text Recognition (HTR) focuses on transforming textual content from images into machine-readable representations. In practice, this enables documents to be stored, searched, and processed more effectively \cite{11303599}.   Beyond these practical applications, HTR can also aid in the preservation of culture and history. An accurate HTR system can help to digitize ancient manuscripts, making these historical documents accessible to the general public and researchers. 

Recent advances in HTR have followed a clear progression, from early rule-based and character-level approaches to statistical models such as Hidden Markov Models, and more recently to deep learning architectures including CNNs, RNNs, and Transformer-based models \cite{11303599}. Much of this development has been reported on Latin-script datasets \cite{11303599}, where early work focused on isolated character recognition, later extending to word- and line-level transcription, and eventually to end-to-end models operating on full text lines and documents \cite{COQUENET2026113373}.

HTR systems for Arabic-script languages have been studied in prior work, with dedicated models and datasets developed for tasks such as word- and line-level recognition. Still, Arabic-script languages have not received as much attention as Latin-script HTR.

The Arabic script is one of the most widely used writing systems in the world. Not only is it the writing system for the Arabic language with over 400 million people speaking Arabic \cite{saeed2024muharaf,rasheed2023multi}, the script is also used for other languages with hundreds of millions of native speakers, including Urdu \cite{maqsood2023unified} with over 200 million speakers, Persian \cite{Wikipedia_2026} with over 100 million speakers, as well as several other languages, such as Pashto, Kurdish, Sindhi, Uyghur, and several African languages written in Ajami. This has resulted in a large and diverse collection of handwritten documents across regions and time periods.

Considering the scale and diversity of Arabic-script documents, robust HTR systems are needed to ensure reliable digitization and to improve access to both historical and modern handwritten materials.


Arabic is written from right to left in a connected script, where characters change shape depending on their position within a word, in contrast to Latin scripts where such shape variation is more limited, mainly appearing as capitalization of initial characters. In addition, diacritics and dot patterns, which occur more frequently than in Latin scripts and introduce further ambiguity, making HTR for Arabic-script languages particularly challenging.

In this work, we focus on line-level HTR, which has become the state-of-the-art (SoTA) in modern HTR systems. Despite recent progress in HTR, several studies \cite{al2026cer,al2026Cross,aljishi2024comparative,chan2024hatformer} have reported that Arabic-script recognition still lags behind Latin-script performance. However, there are limited works that compare multiple datasets across different languages and scripts using a unified setting. \textbf{As a result, it remains unclear how much of this gap is due to script characteristics, data quality, or dataset-specific factors.}

To address this gap, we pose the following research questions:

\textbf{RQ1: What is the performance gap between Arabic-script and Latin-script HTR across datasets and training sizes?} 

\textbf{\textbf{RQ2: What is the trade-off between the number of text lines and the number of characters in achieving comparable recognition performance across scripts, and how are script-specific features involved?}}
 \newline

\noindent \textbf{Contributions.}
This paper makes the following contributions:
\begin{itemize}
    \item We present the first systematic comparison of Arabic-script and Latin-script HTR using a unified model across multiple languages and training sizes.

    \item We quantify the performance gap between Arabic-script and Latin-script HTR, showing that it decreases with more data but persists at full scale. We further compare recognition in terms of the number of text lines and characters, revealing an equivalence trade-off in the data required to achieve comparable CER across scripts.

    \item We provide a character-level analysis of shape variability and error patterns, showing that visually similar character confusions account for a larger proportion of errors in Arabic-script datasets, and that character frequency distributions are significantly more heavy-tailed than in Latin scripts.
    
\end{itemize}

\section{Challenges in Arabic-Script HTR} \label{sec:challenges}
Arabic-script HTR faces challenges due to script properties and data characteristics (modern or historical), including cursive connectivity, contextual character shapes, diacritics, and complex ligatures, which make segmentation and recognition difficult \cite{salaheldin2025advancements}.


An essential challenge in Arabic-script HTR lies in the \textbf{contextual variability of character shapes}. Arabic scripts have different shapes for the same character depending on its position within a word (isolated, initial, medial, and final forms), for example {\arabicfont ,ح ,حـ ,ـحـ ـح}. This increases the number of visual classes that a recognition system must distinguish, as each character can appear in multiple forms \cite{al2026understanding,salaheldin2025advancements}. While Latin scripts also show variations such as uppercase and lowercase forms, these are more limited and do not wholly depend on character position (see Section~\ref{sec:error_analysis}).




Another challenge is the presence of \textbf{diacritics and dots (i‘jām and tashkīl)}. The Arabic script uses dots to distinguish between letters that share the same base shape, and additional diacritical marks to indicate short vowels. These marks are often small, faint, or omitted in handwritten text, making them difficult to detect. While Latin scripts also use diacritics, these are less frequent, whereas in Arabic dots are an integral part of distinguishing between characters.

For example, the letters {\arabicfont ,ح ,خ ج } share the same base shape of the letter {\arabicfont ح} but differ in the use of dots to indicate which specific characters are being represented. These dots are often omitted or misplaced in handwritten text, leading to frequent confusion.

\begin{figure}[h]
    \centering

    \begin{subfigure}[b]{0.30\textwidth}
        \centering
        \includegraphics[height=4cm,width=\textwidth]{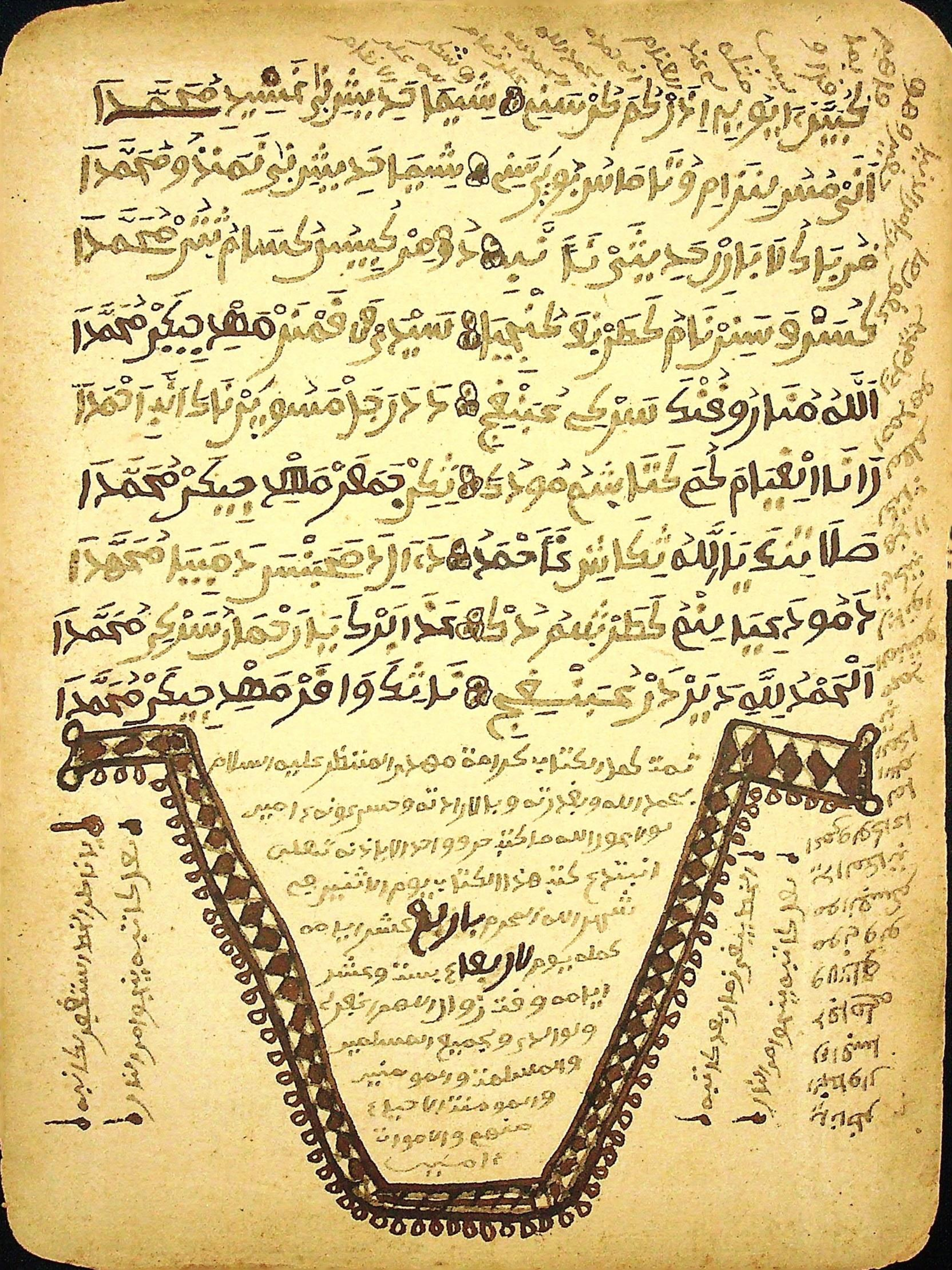}

        \caption{Ajami dataset.}
    \end{subfigure}
    \hfill
    \begin{subfigure}[b]{0.30\textwidth}
        \centering
        \includegraphics[height=4cm,width=\textwidth]{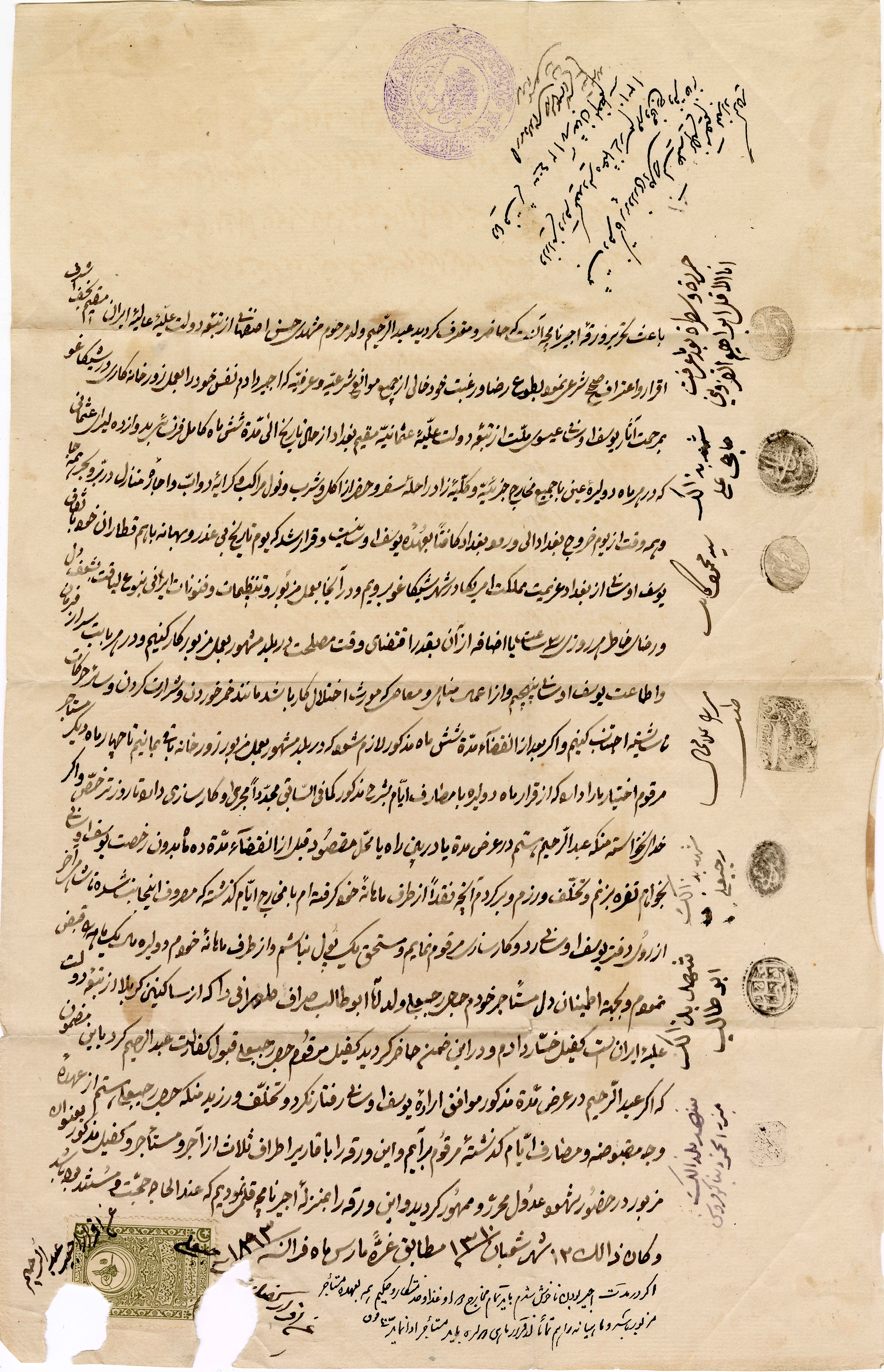}
        \caption{Muharaf dataset.}
    \end{subfigure}
    \hfill
    \begin{subfigure}[b]{0.30\textwidth}
        \centering
        \includegraphics[height=4cm,width=\textwidth]{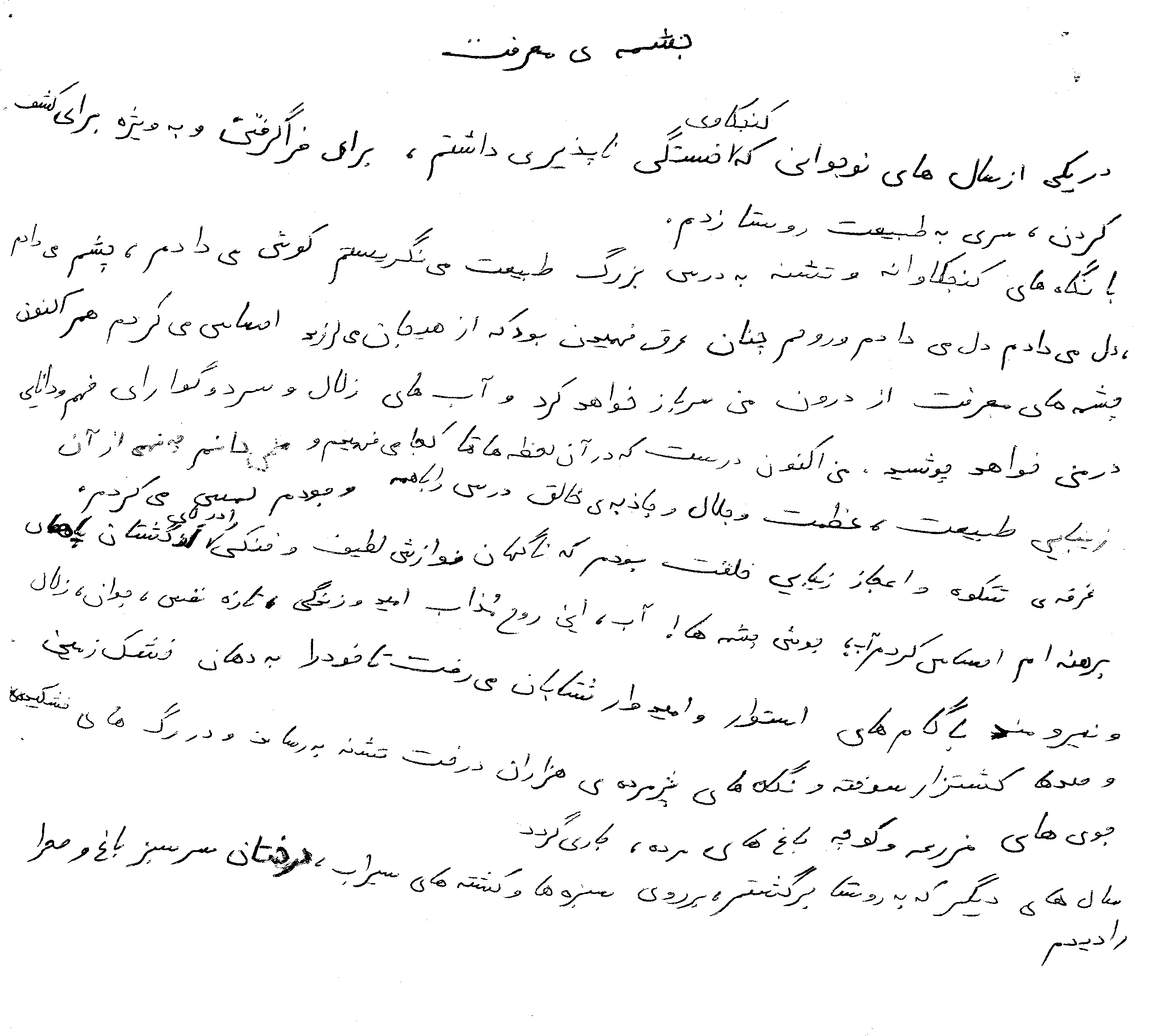}
        \caption{PHTD dataset.}
    \end{subfigure}

\caption{Representative samples from datasets used in this paper.}\label{fig:datasets_samples}
\end{figure}

In addition, the use of diacritics also presents ambiguity for the readers. For example, the phrase without diacritics of ({\arabicfont التشكيل علامات تستخدم العربية اللغة}) becomes ({\arabicfont التَّشْكِيلِ عَلاَمَاتِ تُسْتَخْدِمُ الْعَرَبِيَّةُ اللُّغَةُ}) when the vowels are added. These marks often appear as small marks above or below the characters in the text, and are often confused with dots or are entirely omitted. In this study, one of the Arabic datasets we use, Ajami, contains many such diacritics, as illustrated in Figure~\ref{fig:datasets_samples}.

\textbf{Document degradation and layout variability} also pose challenges, especially for historical manuscripts \cite{saeed2024muharaf}.  As shown in Figure~\ref{fig:datasets_samples}, text lines vary in orientation and alignment, reflecting real-world writing conditions and increasing recognition difficulty.

Finally, a critical limitation in Arabic-script HTR is the \textbf{scarcity of large-scale, diverse annotated datasets}. Compared to Latin-based languages, Arabic-script datasets are limited and less diverse. This scarcity limits the ability of deep learning models to generalize effectively, especially for historical and low-resource settings such as Ajami \cite{chan2024hatformer,saeed2024muharaf,yousuf2025handwritten}.

\section{Methodology} \label{sec:methodology}

This work investigates the performance gap between Arabic-script and Latin-script HTR under controlled experimental conditions. 
To isolate the effects of script and data-related factors, we use a unified setup in which a single recognition model is trained and evaluated across multiple datasets and training sizes, with all experiments conducted at the line level using the provided segmentations and identical model, preprocessing, and training configurations.
To analyze the impact of data availability, we simulate varying resource conditions by training models on subsets of increasing size. Specifically, for each dataset, we use training sizes $K \in \{100, 500, 1000, 2000, 3000, \dots, K_{\text{full}}\}$, where $K_{\text{full}}$ corresponds to the full training set.

\textcolor{black}{The recognition model used in the base experiments is a CRNN \cite{retsinas2022best}} -- a strong and widely used baseline for line-level HTR, which has demonstrated robust performance across diverse datasets \cite{al2026cer,corbille2025applying,imbert2026domain}. Compared to Transformer-based models, CRNNs require less training data and provide a more stable basis for controlled comparisons, making them suitable for analyzing performance differences across scripts under varying training sizes.

The CRNN architecture \cite{retsinas2022best} combines a convolutional feature extractor with recurrent sequence modeling trained using the CTC objective. The convolutional backbone extracts hierarchical visual features from input text-line images, which are converted into a one-dimensional sequence through column-wise pooling along the vertical axis. This sequence is then processed by stacked bidirectional LSTM layers to model contextual dependencies in both directions. The resulting features are projected onto the character space and optimized using the CTC loss, enabling alignment-free training.
In addition, an auxiliary CTC branch is attached directly to the convolutional features during training to improve optimization stability. The architecture details are summarized in Table~\ref{tab:crnn_arch}.

\begin{table}[t]
\centering
\caption{CRNN architecture with CNN feature extractor, BiLSTM layers, and CTC training. An auxiliary CTC branch is used during training.}
\label{tab:crnn_arch}
\resizebox{0.6\linewidth}{!}{
\begin{tabular}{l}
\midrule
\textbf{CRNN Architecture} \\ \midrule
\multicolumn{1}{l}{\textbf{CNN Feature Extractor}} \\
$7\times7$ Conv, 32, stride $2\times2$ \\
Residual Block $\times 2$ (64 channels) \\
\quad each block: $3\times3$ Conv $\rightarrow$ BN $\rightarrow$ ReLU $\rightarrow$ $3\times3$ Conv $\rightarrow$ BN, with identity shortcut \\
Max Pooling ($2\times2$) \\
Residual Block $\times 4$ (128 channels) \\
\quad same structure as above \\
Max Pooling ($2\times2$) \\
Residual Block $\times 4$ (256 channels) \\
\quad same structure as above \\
\multicolumn{1}{l}{\textbf{Sequence Conversion}} \\
Column-wise max pooling (collapse height) \\
\multicolumn{1}{l}{\textbf{Recurrent Sequence Modeling}} \\
BiLSTM (256 units per direction) $\times 3$ \\
\multicolumn{1}{l}{\textbf{Output Layers}} \\
Fully connected layer (character classes) \\
CTC loss (main sequence prediction) \\
Auxiliary CNN projection + CTC loss (training only) \\
\midrule
\end{tabular}
}
\end{table}
\textcolor{black}{To assess whether the observed trends are architecture-dependent, an additional comparison using the HTR-VT Transformer architecture is presented in Section~5.5.
HTR-VT  \cite{li2025htr} uses a modified ResNet-18 feature extractor followed by a 4-layer Vision Transformer encoder with 6 attention heads. Character predictions are generated using a CTC objective, while span masking, Sharpness-Aware Minimization and exponential moving average are employed during training to improve data efficiency and regularization.}

\section{Experiments} \label{sec:experiments}
This section describes the experimental setup used to evaluate the performance gap between Arabic-script and Latin-script HTR. The datasets used in this study are first presented, followed by the implementation details.

\subsection{Datasets}

This study employs nine HTR datasets, including  six Arabic-script and three Latin-script datasets. To ensure comparability across datasets, all experiments are conducted at the line level.

The Arabic-script group comprises Muharaf~\cite{saeed2024muharaf}, a historical Arabic manuscript dataset; PHTI \cite{hussain2022phti}, a large-scale Pashto dataset; the Ajami dataset~\cite{yousuf2025handwritten}, which includes historic West African manuscripts written in Hausa and Fulfulde using Arabic script; PHTD~\cite{alaei2012dataset}, a Persian dataset; KHATT~\cite{mahmoud2014khatt}, a widely used benchmark for modern Arabic handwriting; and NUST-UHWR~\cite{ul2022convolutional}, a modern multi-domain Urdu corpus with contributions from a large number of writers.


\textcolor{black}{For Latin-script comparison, we use NorHAND-mini\footnote{The released split is available at Zenodo: \url{https://zenodo.org/records/20524615}.}, which contains historical Norwegian handwritten documents. 
We restrict training to this a randomly sampled subset of version 3 of the of the original NorHAND dataset~\cite{beyernorhand} which contains more than 200K training lines to obtain a training size comparable to the other datasets used in this study. }
We further use READ-2016~\cite{sanchez2016icfhr2016}, comprising historical German manuscripts, and IAM~\cite{marti2002iam}, a widely used benchmark for modern English handwriting.

\begin{figure}[t]
\centering

\includegraphics[width=0.6\textwidth]{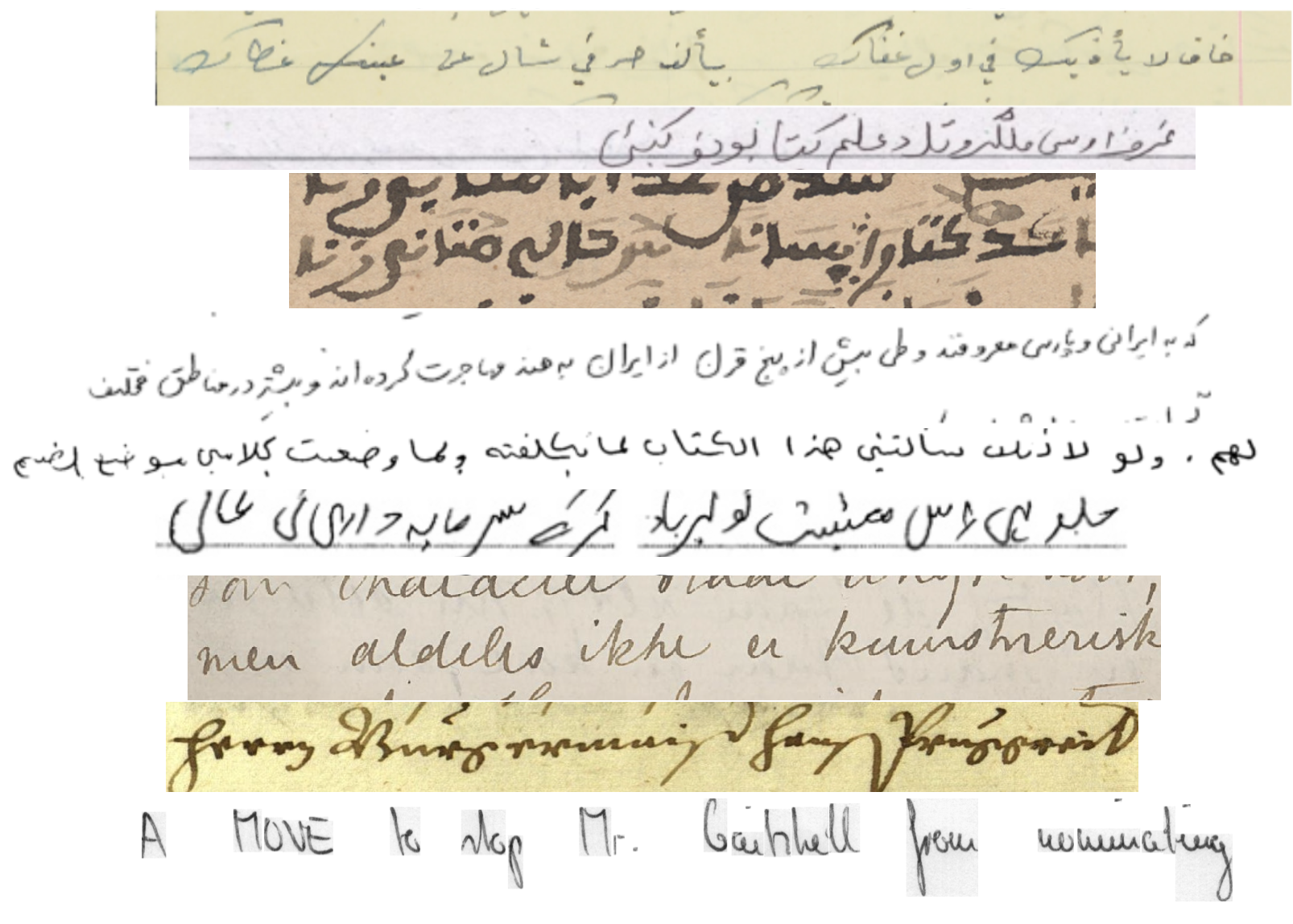}
\caption{Examples of handwriting from the datasets used in this work. From top to bottom: Muharaf (Arabic), PHTI (Pashto), Ajami (Hausa and Fulfulde), PHTD (Persian), KHATT (Arabic), and NUST-UHWR (Urdu). Latin-script datasets include NorHAND (Norwegian), READ-2016 (German), and IAM (English).}
\label{fig:dataset}
\end{figure}

Across all datasets, there is variation in size, writing style, and data quality. Representative samples are shown in Figure~\ref{fig:dataset}, and detailed dataset statistics are provided in Table~\ref{tab:dataset_stats}.
When available, we adopt the standard predefined splits provided with each dataset.
For IAM, we follow the commonly used Aachen split.
For datasets without official splits, such as Ajami and PHTD, we follow the splits used in recent prior work~\cite{al2026cer,al2026Cross}.

\begin{table}[t]
\caption{Dataset statistics. Hist. indicates historical datasets.}
\label{tab:dataset_stats}
\centering
\small
\resizebox{0.8\textwidth}{!}{
\begin{tabular}{l c c c c c}
\midrule
\textbf{Dataset} & \textbf{Language} & \textbf{~Hist.~} & \textbf{Train/Val/Test(lines)} & \textbf{Avg. W--H} & \textbf{Max Len} \\ 
\midrule
\midrule

Muharaf & Arabic & $\checkmark$ & 22091/1069/1334 & 597--60 & 181 \\

PHTI & Pashto & $\times$ & 24919/5332/5339 & 992--76 & 120 \\ 

Ajami & Hausa + Fulfulde & $\checkmark$ & 4346/931/931 & 1289--213 & 106 \\

PHTD & Persian & $\times$ & 1473/160/155 & 1960--180 & 193 \\

KHATT & Arabic & $\times$ & 4791/938/967 & 2074--154 & 132 \\

NUST-UHWR & Urdu & $\times$ & 8479/1061/1056 & 950--64 & 116 \\ 
\midrule
\midrule

NorHAND-mini & Norwegian & $\checkmark$ & 10490/1576/1491 & 1472--133 & 168 \\

READ-2016 & German & $\checkmark$ & 8367/1043/1140 & 962--122 & 43 \\ 

IAM & English & $\times$ & 6482/976/2915 & 1698--124 & 91 \\ \midrule

\end{tabular}
}
\end{table}

\subsection{Implementation details}

All experiments were carried out using PyTorch on an NVIDIA A100-SXM4-40GB GPU, with all models trained under a unified setup to ensure fair comparison across datasets. Input images are converted to grayscale and resized to match the average line height of each dataset, as reported in Table~\ref{tab:dataset_stats}, while preserving aspect ratio.  To account for handwriting variability, we apply standard geometric and photometric data augmentations uniformly across datasets.
For Arabic-script datasets, ground-truth transcriptions are reversed during training to align with the left-to-right feature extraction and CTC decoding.

Performance is evaluated using Character Error Rate (CER). For training, models are optimized using AdamW with a batch size of 16 and an initial learning rate of $5\times10^{-4}$. The learning rate was reduced by a factor of 0.1 at 50\% and 75\% of the training process. For each training size $K \in \{100, 500, 1000, 2000,\dots, K_{\text{full}}\}$, models were trained for a fixed number of iterations, and the best checkpoint is selected based on validation CER.

\textcolor{black}{For the HTR-VT~\cite{li2025htr} we follow the original implementation. Images are resized to a fixed input resolution of $512\times64$ while preserving aspect ratio. The model is trained using AdamW with Sharpness-Aware Minimization (SAM), exponential moving average (EMA), cosine warmup scheduling, and span masking. We use a batch size of 128 and train for 100,000 iterations with a maximum learning rate of $10^{-3}$, following the settings reported in~\cite{li2025htr}.}

\section{Results} \label{sec:results}
We first analyze the performance differences between Arabic-script and Latin-script datasets across training sizes. We then study the effect of data quality by comparing results on original and cleaned datasets. Next, we examine the relationship between training data size and character distribution. We analyze errors to identify common patterns and the impact of visually similar characters. We concludeby comparing results with the CRNN with those from the HTR-VT.

\subsection{Performance Gap Between Arabic- and Latin-Script HTR}

\begin{figure}[t]
\centering
\includegraphics[width=\textwidth]{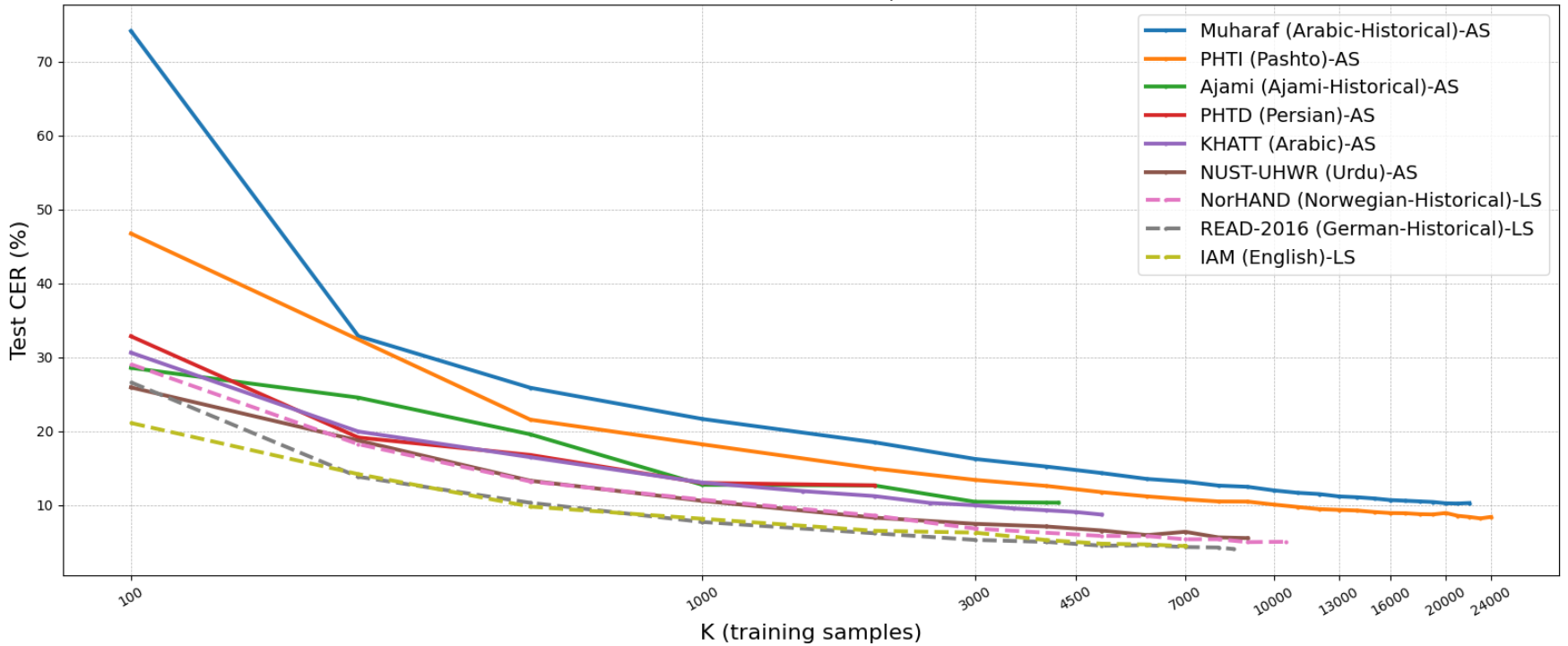}
\caption{Performance gap between Arabic-script (AS) and Latin-script (LS) HTR systems across multiple datasets and data sizes. \textcolor{black}{Solid lines denote Arabic-script datasets and dashed lines denote Latin-script datasets.}}
\label{fig:gap}
\end{figure}

Using the CRNN described in Section~\ref{sec:methodology}, we perform line-level HTR on all nine datasets at various $K$ levels. The results are summarized in Figure~\ref{fig:gap} and Table~\ref{tab:dataset_results}. There is a clear performance gap between Arabic-script and Latin-script datasets across most training sizes with higher CER values for Arabic-script datasets compared to Latin-script datasets across training sizes. An exception is NUST-UHWR which at lower training sizes ($K \leq 2000$) achieves lower CER than some Latin-script datasets.
These trends are consistent with prior observations in the literature, but have not been systematically analyzed under a unified experimental setup across multiple datasets and training sizes.

Among Arabic datasets, the highest error rates are observed for Muharaf, PHTI, Ajami, and PHTD. For Muharaf, PHTI, and Ajami, this is largely due to label errors, including transcription, segmentation, and orientation mistakes, in these datasets. Previous work~\cite{al2026cer} reports that Ajami contains around 8.5\% label errors, while Muharaf exhibits 6--9\% label errors across splits.

In contrast, the weaker performance of PHTD is mainly due to limited data diversity. Although it contains 1,473 training lines, it includes only 738 unique words compared to 19,639 in KHATT, restricting generalization ability.

NUST-UHWR achieves the best performance among Arabic datasets, resulting in the smallest gap compared to Latin-script benchmarks, and for $K < 2000$ it achieves lower CER than some Latin-script datasets (e.g., NorHand). This is explained by its relatively clean data. However, a noticeable gap still remains. For example, at full scale, NUST-UHWR achieves 5.55\% CER, compared to 4.42\% for IAM, showing that the gap persists even under favorable conditions.

Table~\ref{tab:script_gap} summarizes the average performance across scripts, showing a consistent gap between Arabic and Latin datasets across training sizes. The gap is largest in low-resource settings (K=100), decreases with more data, but remains around 5--7 CER points even at full scale, stabilizing beyond K=4000.

We also explored representing Arabic characters using explicit positional forms (e.g., initial, medial, and final) in the training labels to better model Arabic script variability. However, this did not lead to consistent improvements in recognition performance.



\begin{table}[t]
\centering
\caption{Performance gap between AS and LS HTR across datasets and training sizes. \textcolor{black}{Entries marked with ``--'' indicate that the corresponding training size exceeds the number of available training samples in the dataset.}}
\label{tab:dataset_results}
\small
\resizebox{0.8\textwidth}{!}{
\begin{tabular}{l c c c c c c c c}
\toprule
Dataset & Lang & K=100 & K=1000 & K=2000 & K=3000 & K=4000 & K=5000 & Full \\
\midrule
\midrule

Muharaf & Arabic & 74.12 & 21.65 & 18.50 & 16.25 & 15.22 & 14.34 & \textbf{10.30 (K=20667)} \\
PHTI & Pashto & 46.72 & 18.23 & 14.96 & 13.41 & 12.60 & 11.76 & \textbf{8.40 (K=24919)} \\
Ajami & Arabic & 26.62 & 12.75 & 12.66 & 10.45 & 10.35 & -- & \textbf{10.33 (K=4346)} \\
PHTD & Persian & 32.83 & 13.00 & -- & -- & -- & -- & \textbf{12.68 (K=1473)} \\
KHATT & Arabic & 30.63 & 13.06 & 11.23 & 9.99 & 9.31 & -- & \textbf{8.72 (K=4791)} \\
NUST-UHWR & Urdu & 25.92 & 10.17 & 8.31 & 7.48 & 7.11 & 6.58 & \textbf{5.55 (K=8479)} \\

\midrule
\midrule

NorHAND & Norwegian & 29.05 & 10.75 & 8.59 & 6.83 & 5.39 & 5.84 & \textbf{5.05 (K=10490)} \\
READ-2016 & German & 26.61 & 7.73 & 6.20 & 5.30 & 5.04 & 4.52 & \textbf{4.00 (K=8367)} \\
IAM & English & 21.11 & 8.16 & 6.54 & 6.29 & 5.28 & 4.79 & \textbf{4.42 (K=6482)} \\

\bottomrule
\end{tabular}
}

\end{table}

\begin{table}[t]
\caption{Average CER comparison between AS and LS datasets.}
\fontsize{8}{8}\selectfont
\label{tab:script_gap}
\centering

\begin{tabular}{c cc c}
\toprule
 & \multicolumn{2}{c}{Average CER (\%)} &  \\
\cmidrule(lr){2-3}
K & Arabic & Latin & Gap \\
\midrule
\midrule
100  & 52.14 & 25.59 & +26.55 \\
1000 & 18.76 & 8.88  & +9.88 \\
2000 & 14.38 & 7.11  & +7.27 \\
3000 & 12.79 & 6.14  & +6.65 \\
4000 & 12.16 & 5.16  & +7.00 \\
5000 & 10.89 & 5.05  & +5.84 \\ 
Full & 10.23 & 4.46  & +5.77 \\
\bottomrule
\end{tabular}
\end{table}

\subsection{Effect of Data Quality on the Arabic–Latin Performance Gap}
\begin{figure}[t]
\centering
\includegraphics[width=1\textwidth]{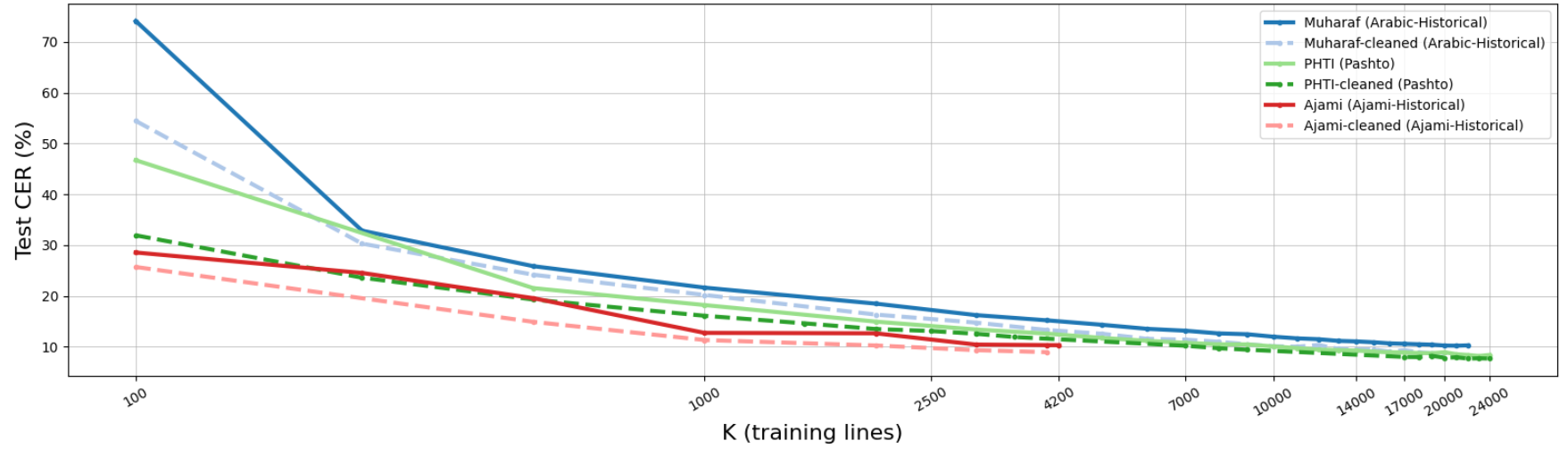}
\caption{Recognition performance for cleaned and non-cleaned datasets across different training sizes for Arabic-script HTR.}
\label{fig:clean}
\end{figure}

\textcolor{black}{It has been observed in prior work~\cite{al2026cer} that the Muharaf, PHTI, and Ajami datasets contain several label errors. Similar issues have not been reported for the Urdu or Latin-script datasets. We therefore compare HTR performance using the original datasets and the publicly released cleaned dataset versions provided by~\cite{al2026cer}.} 
Figure~\ref{fig:clean} compares performance when trained on $K=100$ to the full training set size up to 20,000.
Cleaning consistently improves performance across all datasets, especially in low-resource settings, showing that data quality plays an important role in HTR performance. Cleaning also reduces the gap between Arabic and Latin scripts. For example, at $K=100$, the gap decreases from 24.23 to 11.78 CER points. However, even after cleaning, a noticeable gap remains across all training sizes (Table~\ref{tab:clean_gap_or}).

This indicates that while label errors contributes to the performance gap, it does not fully explain it. This is partly due to properties of the Arabic script, such as positional character variation, which increases the number of distinct visual forms and reduces effective data coverage. A more detailed analysis of these effects is provided in Section~\ref{sec:error_analysis}.

\begin{table}[t]
\centering
\fontsize{8}{8}\selectfont

\caption{
Comparison of average CER between AS and LS datasets before and after data cleaning. 
AS results are computed over paired datasets (Muharaf, Ajami, and PHTI).}
\label{tab:clean_gap_or}
\resizebox{0.65\textwidth}{!}{%
\begin{tabular}{c c c c c c}
\toprule
K & AS (orig) & AS (clean) & LS & Gap (orig) & Gap (clean) \\
\midrule
\midrule
100  & 49.82 & 37.37 & 25.59 & +24.23 & +11.78 \\
1000 & 17.49 & 15.23 & 8.88  & +8.61  & +6.35  \\
2000 & 15.06 & 13.38 & 7.11  & +7.95  & +6.27  \\
3000 & 13.35 & 12.60 & 6.14  & +7.21  & +6.46  \\
4000 & 12.69 & 11.95 & 5.16  & +7.53  & +6.79  \\
Full & 10.43 & 8.83  & 4.46  & +5.97  & +4.37  \\
\bottomrule
\end{tabular}
}
\end{table}

\subsection{Impact of Training Data Size and Character Distribution}
\begin{figure}[t]
\centering
\includegraphics[width=0.75\textwidth]{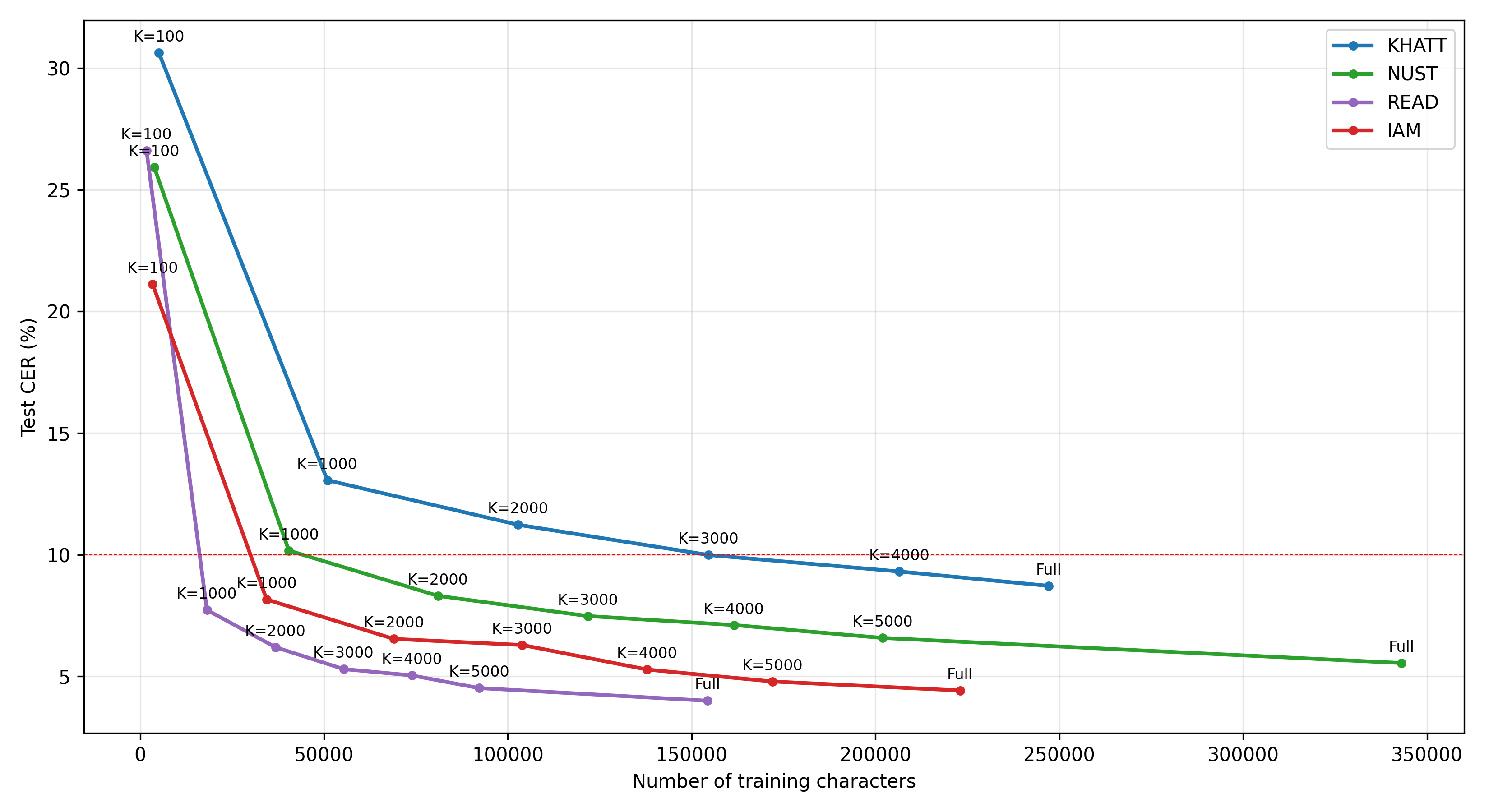}
\caption{Character-level training coverage versus recognition performance across training sizes. The dashed horizontal line marks CER = 10\%, showing how many training characters are needed to reach a comparable performance level across datasets.}
\label{fig:cer_vs_characters}
\end{figure}

The number of characters per line varies across datasets, so the total number of training characters does not scale uniformly with the number of training samples (text lines).
Figure~\ref{fig:cer_vs_characters} shows the relationship between the number of training characters and recognition performance across four datasets. 

As the amount of training text increases, CER decreases for all datasets, but the rate of improvement differs substantially. The dashed line at CER = 10\% highlights this gap more clearly: KHATT needs roughly 150{,}000 training characters to reach this level, whereas IAM reaches comparable performance with only about 34{,}000 characters. This suggests that the same number of training characters does not provide the same impact across datasets. The difference is likely related not only to dataset size, but also to how characters and shapes are distributed within each dataset.

Arabic datasets such as KHATT and Muharaf contain 82 shapes, compared to 50 in IAM and 53 in READ, even when accounting for variations such as uppercase and lowercase letters in Latin-script datasets. As a result, for a fixed number of training samples, each visual pattern is observed less frequently in Arabic, leading to lower effective coverage than in Latin scripts for the same number of training samples. 

\begin{figure}[t]
\centering

\begin{subfigure}[t]{0.85\textwidth}
    \centering
    \includegraphics[width=\linewidth]{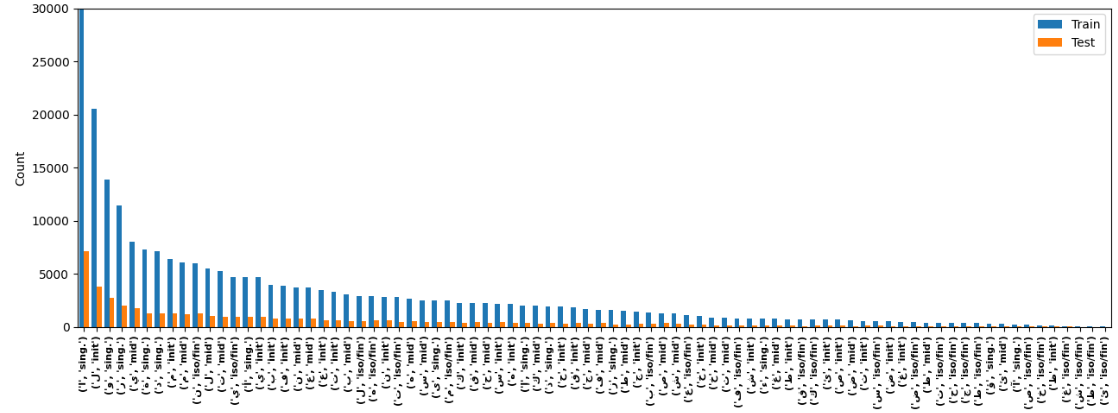}
     \vspace{-5mm}
    \caption{KHATT}
    \label{fig:khatt_shape_dist}
\end{subfigure}


\begin{subfigure}[t]{0.85\textwidth}
    \centering
    \includegraphics[width=\linewidth]{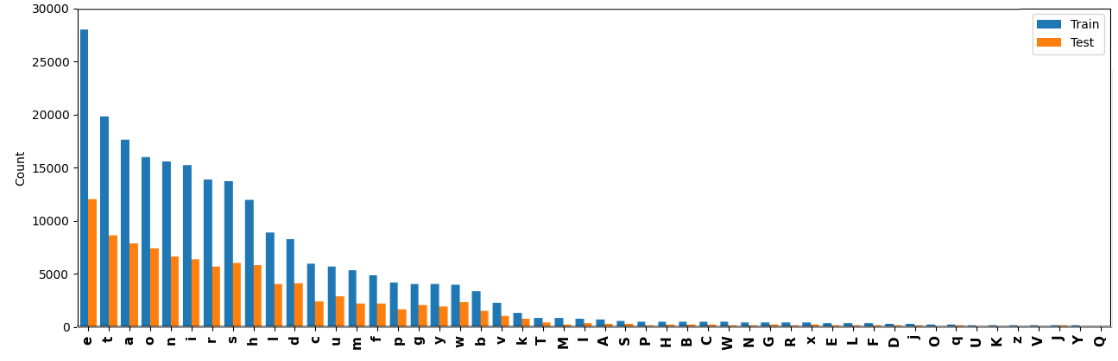}
    \vspace{-5mm}
    \caption{IAM}
    \label{fig:iam_shape_dist}
\end{subfigure}

\caption{Training and test shape-frequency distributions for KHATT and IAM. Arabic shape labels include positional indicators: init (initial), mid (medial), iso/fin (isolated/final), and sing. (single form).}


\label{fig:shape_distribution}
\end{figure}

To better understand this difference, we examine how training data is distributed across character shapes. Figure~\ref{fig:shape_distribution} compares KHATT and IAM. The rare characters in IAM are the uppercase letters and infrequent lowercase letters (e.g., q, x, z). The distribution shows a rapid drop from a small set of very frequent characters to much lower-frequency ones. In contrast, KHATT exhibits a much heavier tail, with many character shapes appearing at relatively low frequencies. This leads to a more dispersed distribution, where training examples are spread over a larger number of visual patterns, reducing effective coverage per shape. Similar character distribution patterns are observed across the other datasets.

While it will take more lines to get enough Latin upper case samples to learn them, errors classifying them will also be infrequent and have a lower impact on total CER. The rapid drop-off in Latin scripts and the heavy-tailed distribution in Arabic scripts are key observations highlighted by this study. 

\subsection{Error Analysis} \label{sec:error_analysis}

We analyze character-level errors by aligning predicted text and ground-truth sequences and classifying them into substitutions, deletions, and insertions. We analyze character-level errors across representative datasets from both Arabic and Latin scripts, including modern and historical collections.

To better understand these errors, we group characters based on visual similarity. We consider two types: 
(i) dot-based groups, where characters share the same base shape but differ in dots (e.g.,{\arabicfont ب–ت–ث, ج–ح–خ, ف–ق}), and 
(ii) shape-based groups, where characters have very similar forms (e.g.,{\arabicfont ت–ن}). For comparison, similar confusions also occur in Latin scripts, such as between characters with similar shapes (e.g., c–e-o, m–n, t–l, u–v).

\begin{table}[t]
\centering
\small

\caption{Error distribution across datasets, including deletion (Del), substitution (Sub), insertion (Ins), and similarity-based substitutions.}

\label{tab:error_types}
\resizebox{0.6\textwidth}{!}{%
\begin{tabular}{lcccc}
\toprule
Dataset & Del & Sub & Ins & Similarity \\
\midrule
\midrule
Muharaf (Arabic) 
& ~~2587 (4.70\%)~ 
& ~2292 (4.17\%)~ 
& ~927 (1.69\%)~ 
& ~674 (29.4\%)~ \\
KHATT (Arabic)   
& 743 (1.56\%) 
& 2354 (4.93\%) 
& 387 (0.81\%) 
& 700 (29.7\%) \\

\midrule
\midrule

READ (German)    
& 185 (0.99\%) 
& 392 (2.10\%) 
& 95 (0.51\%) 
& 97 (24.7\%) \\

IAM (English)    
& 832 (0.85\%) 
& 3322 (3.39\%) 
& 604 (0.62\%) 
& 510 (15.4\%) \\

\bottomrule
\end{tabular}
}
\end{table}
Table~\ref{tab:error_types} summarizes the normalized distribution of error types across Arabic- and Latin-script datasets. Across Arabic datasets, substitution errors account for a larger proportion of total errors compared to Latin datasets. Furthermore, a substantial portion of these substitutions is attributed to visually similar characters, with approximately 29--30\% in Arabic datasets compared to 24.7\% in READ and 15.4\% in IAM.

Given that visually similar character confusions account for a large proportion of errors in Arabic datasets ($\thickapprox30\%$), we further analyze these patterns using representative modern and historical datasets (KHATT and Muharaf).
\begin{figure}[t]
\centering

\begin{subfigure}[t]{0.48\textwidth}
    \centering
    \includegraphics[width=0.9\linewidth]{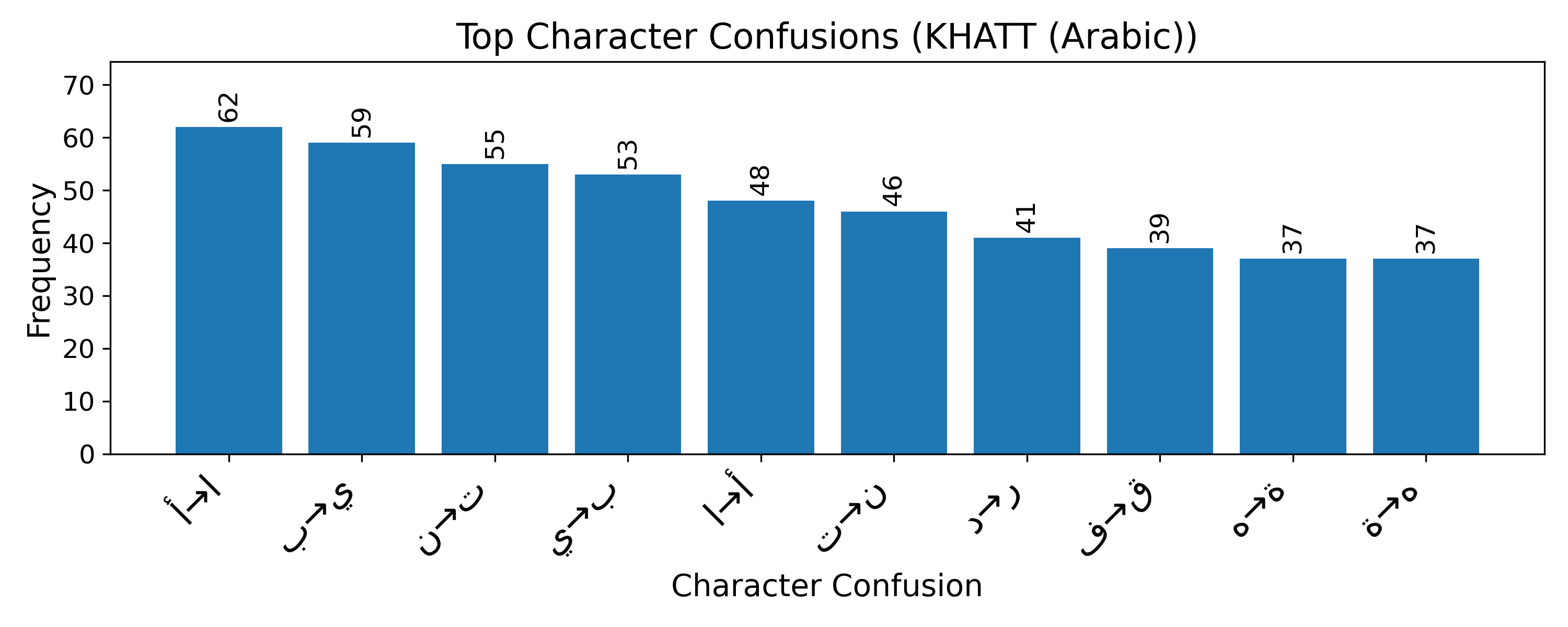}
    \caption{KHATT}
\end{subfigure}
\hfill
\begin{subfigure}[t]{0.48\textwidth}
    \centering
    \includegraphics[width=0.9\linewidth]{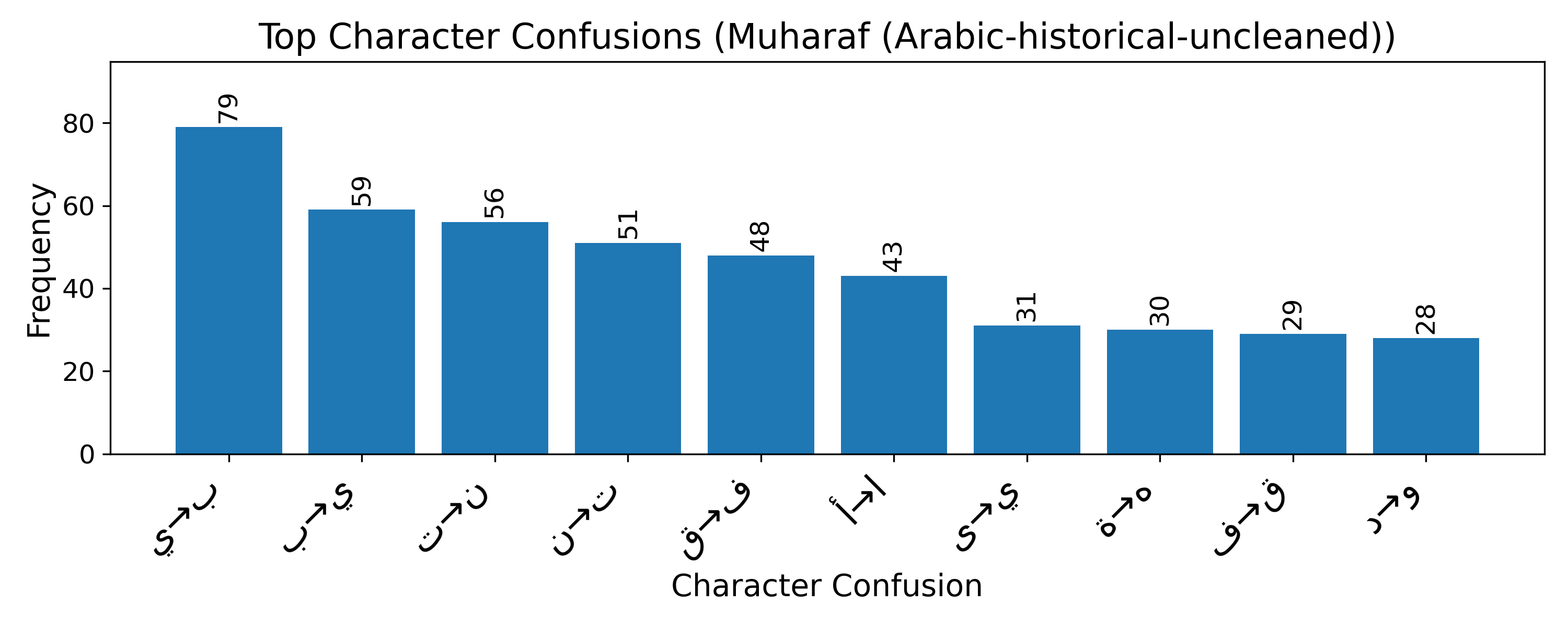}
    \caption{Muharaf}
\end{subfigure}

\caption{Top character confusions for Arabic datasets. Errors are dominated by substitutions between visually similar characters.}

\label{fig:error_analysis}
\end{figure}
Figure~\ref{fig:error_analysis} shows that the most common errors involve characters with similar shapes or dot patterns (e.g., {\arabicfont ف $\leftrightarrow$ ق ,
ب $\leftrightarrow$ ي ,
ت  $\leftrightarrow$ ن}).







\subsection{Architecture Comparison}

\textcolor{black}{To assess whether the observed performance gap between Arabic-script (AS) and Latin-script (LS) datasets is primarily driven by model architecture, we conducted a controlled comparison between our CRNN baseline \cite{retsinas2022best} and HTR-VT \cite{li2025htr}, a Transformer-based handwritten text recognition model. Table~\ref{tab:arch_compare} reports the resulting CER.
The results show that the relative difficulty of Arabic-script datasets persists across both architectures. While performance varies between datasets, both CRNN and HTR-VT achieve substantially lower CER on the Latin-script IAM dataset than on Arabic-script datasets such as KHATT and Muharaf. For example, HTR-VT achieves a CER of 4.78\% on IAM, compared with 8.91\% on KHATT and 11.79\% on Muharaf. 
Although HTR-VT contains substantially more parameters (53.5M vs. 10M) and requires longer training times, its performance remains comparable to the CRNN. 
}
\begin{table}[t]
\centering
\fontsize{8}{8}\selectfont
\caption{Comparison of representative architectures across Arabic- and Latin-script datasets. External results are not directly comparable due to differing training setups, but illustrate typical performance ranges.}
\label{tab:arch_compare}
\renewcommand{\arraystretch}{1.05}

\resizebox{0.6\textwidth}{!}{%
\begin{tabular}{l l c c}
\toprule
\textbf{Dataset (Language)} & \textbf{Model} & \textbf{Type} & \textbf{CER (\%)} \\
\midrule

\multirow{5}{*}
{\shortstack[l]{KHATT\\(Arabic)}}
 & CRNN (Ours) & CRNN-based & \textbf{8.45} \\
 \cdashline{2-4}
 & HTR-VT (Ours) & Transformer & 8.91 \\
 & DAN \cite{aljishi2024comparative} & Transformer & 8.90 \\
 & TrOCR \cite{chan2024hatformer} & Transformer & 15.40 \\
\midrule

 \multirow{3}{*}
{\shortstack[l]{Muharaf\\(Arabic)}}
 & CRNN (Ours) & CRNN-based & \textbf{10.11} \\
 \cdashline{2-4}
 & HTR-VT (Ours) & Transformer & 11.79 \\
 & TrOCR \cite{chan2024hatformer} & Transformer & 11.70 \\
\midrule

 \multirow{3}{*}
{\shortstack[l]{NUST-UHWR\\(Urdu)}}
 & CRNN (Ours) & CRNN-based & 5.55 \\
 \cdashline{2-4}
 & HTR-VT (Ours) & Transformer & 6.12 \\
 & Conv-Transformer \cite{riaz2022conv} & Transformer & \textbf{5.31} \\
\midrule

 \multirow{4}{*}
{\shortstack[l]{IAM\\(English)}}
 & CRNN (Ours) & CRNN-based & 4.42 \\
 & CRNN+Center loss \cite{corbille2025applying} & CRNN-based & 3.91 \\
 \cdashline{2-4}
 & HTR-VT (Ours) & Transformer & 4.23 \\
 & TrOCR Large \cite{li2023trocr} & Transformer & \textbf{2.89} \\
\midrule

READ & CRNN (Ours) & CRNN-based & \textbf{4.00} \\
 \cdashline{2-4}
(German) & HTR-VT (Ours) & Transformer & 4.38 \\
\midrule

\end{tabular}
}
\end{table}

\section{Conclusion} \label{secconclusion}

Our results demonstrate that Arabic-script datasets consistently show higher CERs, with a large gap in low-resource settings. Data quality plays an important role, as cleaning reduces error rates and narrows the gap. However, a consistent gap remains after controlling for data quality and scale, indicating that properties of the Arabic script—such as contextual variation also contribute to the difficulty. Our error analysis shows that visually similar character confusions account for a larger proportion of errors in Arabic-script datasets (around 30\%) than in Latin-script datasets (15–25\%). Character frequency analysis shows that Arabic has a more heavy-tailed distribution, so many characters receive fewer training examples, reducing effective coverage per shape. More training data is therefore required to achieve the same CER, and comparisons at equal $K$ lines are less appropriate. Larger clean Arabic datasets are needed to address this limitation.
Future work should consider synthetic datasets to better isolate data quality and script effects.

\section{Acknowledgements} 
This work was partially supported by the Wallenberg AI, Autonomous Systems and Software Program (WASP) and financially supported by the European Regional Development Fund and the MARTINA-project (no. 20367152).

%
%
%
%
\bibliography{ref}
\bibliographystyle{plain}

\end{document}